# Global solar irradiation prediction using a multi-gene genetic programming approach

Indranil Pan[a,b*], Daya Shankar Pandey[a], Saptarshi Das[c,d]

a) Centre for Energy Studies, Indian Institute of Technology Delhi, Hauz Khas, New Delhi 110 016, India.
b) Energy, Environment, Modelling and Minerals ($E^2M^2$) Research Section, Department of Earth Science and Engineering, Imperial College London, Exhibition Road, London SW7 2AZ, United Kingdom.
c) Department of Power Engineering, Jadavpur University, Salt Lake Campus, LB-8, Sector 3, Kolkata 700098, India.
d) Communications, Signal Processing and Control (CSPC) Group, School of Electronics and Computer Science, University of Southampton, Southampton SO17 1BJ, United Kingdom.

Emails: indranil.jj@student.iitd.ac.in, i.pan11@imperial.ac.uk (I. Pan*),
dspiitd2010@gmail.com (D.S. Pandey)
saptarshi@pe.jusl.ac.in, s.das@soton.ac.uk, (S. Das)

**Abstract:** In this paper, a nonlinear symbolic regression technique using an evolutionary algorithm known as multi-gene genetic programming (MGGP) is applied for a data-driven modelling between the dependent and the independent variables. The technique is applied for modelling the measured global solar irradiation and validated through numerical simulations. The proposed modelling technique shows improved results over the fuzzy logic and artificial neural network (ANN) based approaches as attempted by contemporary researchers. The method proposed here results in nonlinear analytical expressions, unlike those with neural networks which is essentially a black box modelling approach. This additional flexibility is an advantage from the modelling perspective and helps to discern the important variables which affect the prediction. Due to the evolutionary nature of the algorithm, it is able to get out of local minima and converge to a global optimum unlike the back-propagation (BP) algorithm used for training neural networks. This results in a better percentage fit than the ones obtained using neural networks by contemporary researchers. Also a hold-out cross validation is done on the obtained genetic programming (GP) results which show that the results generalize well to new data and do not over-fit the training samples. The multi-gene GP results are compared with those, obtained using its single-gene version and also the same with four classical regression models in order to show the effectiveness of the adopted approach.

1. Introduction

Incoming global solar radiation on earth's surface is an essential input parameter for many applications viz. design of solar energy systems, architectural design, forecasting weather conditions, solar water heating system, crop growth etc. However collecting and storing solar radiation data over an entire country like India is challenging. Due to scarcity of available measured data at remote or rural locations, theoretical available solar energy can be predicted from the existing meteorological data. It is an important issue to estimate solar energy on earth surface to design a mechanism to utilize solar energy effectively.

The Jawaharlal Nehru National Solar Mission is an ambitious initiative taken by the Government of India together with state governments to tackle the issues of energy security in India and foster the growth of renewable energy. It shows India's concern towards sustainable energy security and is an initiative towards mitigating climate change. Being a tropical country, India receives almost 5000 trillion kWh energy from the sun per year which is much higher than the total energy consumption of the whole nation. However, the share of solar energy is a merely 0.4% against other sources of energy production in India [1]. Therefore, viability analysis of solar energy and estimation of the solar energy potential at different locations of the country is essential. Therefore, in the last few years numerous mathematical and statistical models and others based on fuzzy logic, artificial neural networks, particle swarm optimization (PSO) etc. have been developed over time for estimating available solar irradiance from the existing measured parameters such as air temperature, relative humidity, sunshine duration, cloudiness etc. [2].

Most of the statistical or regression based models assume an underlying template for the prediction formula and use some form of mean squared error (MSE) minimization technique to arrive at a fit for the measured data. This technique depends more on the insight of the design engineer and choosing the appropriate model becomes very heuristic and is not streamlined. Moreover, it is not easy to heuristically propose a highly complex nonlinear model which might be very good at solar irradiation prediction. To reduce these hidden heuristics, the symbolic regression method by Genetic Programming is a viable alternative [3]. Even though GP is meta-heuristic in nature, the parameters to be chosen in a GP run, like the number of generations, set of analytical functions, cross-over & mutation rates etc. have a cause-effect relationship on the final results which can be quantified to some extent. The effects of these parameters on the final output have been studied for many test bench problems and it is known for example, that to obtain good solutions it is important to have a much higher value of crossover fraction than the mutation fraction. Such kind of cause-effect relationship of the GP parameters on the final evolved solution narrows down the meta-heuristics to a certain extent. For a person with adequate background in GP algorithms, it then takes only a few iterations to arrive at a good result, even when applying the algorithm to a completely new field. However on the other hand, choosing a nonlinear model from scratch becomes very application specific. The designer needs to have much insight in the dataset and the actual physics behind the problem to propose a nonlinear model. Hence the symbolic regression method by genetic programming is proposed as a viable alternative [3]. GP and other meta-heuristic techniques have been successfully applied in similar real world problems as discussed in [4], [5].

GP based regression techniques have a flexible template and can evolve the structure of the prediction equation based on the mathematical operators supplied by the designer, while trying to minimize an objective function (the prediction error in this case). Other models which use fuzzy logic or neural networks are good at approximating a particular dataset and have high prediction accuracies. However, these are black-box models and need sophisticated software for deployment to be used by end users. Hence it is not suitable for the end user who might be a field engineer trying to arrive at solar irradiance calculation through a simple hand held calculator. This disadvantage is also removed by the GP technique since it results in analytical expressions which can be hand calculated.

A new variant of GP known as the multi-gene GP or MGGP is used in the present study and also compared with the results from the traditional variant i.e. single gene GP (SGGP). It has been illustrated in [6] that this multi-gene GP methodology outperforms the traditional single gene GP on some benchmark problems and is more expedient in prediction with lesser number of terms. Application of MGGP algorithm in real world engineering problems can be seen in [7], [8]. Also, hybridization of GP and simulated annealing has been applied for solar irradiation prediction in Mostafavi *et al.* [9]. The measured meteorological data by Indian Meteorological Department (IMD) Pune, compiled by Mani [10] is used in the present study to make a GP based model for the forecast of solar irradiation. It includes data such as monthly mean solar radiation on horizontal surface, mean duration sunshine per hour, height above sea level for these stations.

The main objective of the present study is to show an application of MGGP approach in predicting global solar irradiation. To the best of the authors' knowledge this is the first study using MGGP approach to predict global solar irradiation. Simulation comparisons with a simpler SGGP algorithm and standard regression models have also been presented.

The rest of the paper is organized as follows. Section 2 presents a brief overview of the different techniques employed by contemporary researchers in the recent past and the achieved accuracies in prediction with these approaches. Section 3 discusses the Genetic Programming method for symbolic regression. Section 4 talks about the nature of the data used in the simulation study along with the basic terminology of the prediction variables. The results with the MGGP algorithm are illustrated in Section 5 along with a few discussions. Simulation comparison of the MGGP predictions with that using relatively simpler SGGP method and standard regression analyses are discussed in section 6. The paper ends in Section 7 with the conclusions followed by the references.

2. **Literature review and model comparison**

In recent literatures several adaptive models are used e.g. Mellit *et al.* [11] used an adaptive ANN algorithm to optimize the size of stand-alone photovoltaic systems where the model was combined with radial basis function (RBF) type NN and infinite impulse response (IIR) filter. Mellit *et al.* [12] developed a hybrid model combined with ANN and a library of Markov transition matrices to predict global solar radiation data in Algeria using an input data latitude, longitude and altitude. Mellit *et al.* [13] used an adaptive wavelet-network model for predicting total solar radiation and compared the result with other technique viz. AutoRegressive (AR), AutoRegressive Moving Average (ARMA), Multi-Layer Perceptron (MLP) etc. Mellit and co-researchers [14] illustrated an adaptive α-model and Feed-Forward Neural Network (FFNN) to forecast hourly global, diffuse and direct solar irradiance using sunshine hours, air temperature and relative humidity where it is concluded that the FFNN prediction is better than the adaptive α-model in forecasting solar radiation.

Among several evolutionary and swarm algorithms the application of PSO has been popular in the field of solar energy. Behrang *et al.* [15] used PSO algorithm to obtain Angstrom coefficient and developed five new models to predict global solar radiation on horizontal surface in Iran. Reikard [16] carried out a comparative study between AutoRegressive Integrated Moving Average (ARIMA), neural network and hybrid model for predicting solar radiation. Mohandes [17] used an improved ANN technique with PSO by making a comparison between PSO-ANN and BP-ANN algorithm to predict global solar radiation.

Application of approximate inference or fuzzy logic has also been there in solar irradiation prediction. Sen [18] has found that fuzzy algorithm estimate solar radiation from sunshine duration more precisely as compared to Angstrom linear regression method. However, it has ignored certain meteorological factor data viz. temperature, altitude, elevation, wind velocity etc. These factors are associated to estimated mean daily global solar radiation on horizontal surface (H) and mean sunshine duration (S) and their negligence initiates an error whilst estimating solar radiation. A methodology of predicting hourly solar radiation has been proposed by Sfetsos and Coonick [19] using ANN and Adaptive Neuro-Fuzzy Inference System (ANFIS) technique. The authors have used temperature, pressure, air speed and direction as input parameters while predicting hourly solar radiation. Inclusion of certain additional meteorological data can boost forecasting capability of the proposed model. An attempt was made in [20] to implement fuzzy set theory to estimate solar irradiation using air temperature as an input parameter. Most recently, Boata and Gravila [21] have proposed a new fuzzy model for forecasting stochastically global solar irradiation.

It has been seen in recent years that various paradigms of AI techniques, especially several ANN variants are applied to solve solar energy related modelling and prediction problems. Computing global solar radiation using empirical method requires a set of equations which correlates with meteorological data. For that reason, an alternative method has to be developed for estimating these data at the different locations where measured data are not available. The application of ANN is a powerful technique used for estimating global solar radiation. It pays attention to the important inputs and ignores insignificant or uncorrelated excess data [22]. This technique is capable of approximating any continuous linear functions with an arbitrary accuracy. ANN prediction is based on prior available data and is therefore commonly preferred by the researchers over other theoretical and empirical methods. Mohandes *et al.* [23] presented a neural network technique to predict global solar radiation in the Kingdom of Saudi Arabia using latitude, longitude, altitude and sunshine duration as input parameters. Mohandes *et al.* [24] have also used RBF technique for modelling global solar radiation using latitude, longitude, altitude and sunshine duration as input parameters. Reddy and Ranjan [25] presented an ANN algorithm based model for estimation of monthly, daily and hourly values of global solar radiation at 13 different stations (six from South India and five from North India) demonstrating its dominance over classical regression methods. Benghanem *et al.* [26] developed six ANN models for predicting global solar radiation altering combinations of input parameters. Alam *et al.* [27] developed a feed forward ANN model using back propagation algorithm for estimating monthly, mean hourly and daily diffuse solar radiation at different climatic stations in India. Behrang *et al.* [28] considered the effect of different meteorological input parameters and proposed six ANN models to predict global solar radiation in Iran based on MLP and RBFNN. Abdulazeez [29] have applied ANN technique to predict monthly solar radiation using sunshine hours, maximum ambient temperature and relative humidity for Nigeria. Yadav and Chandel [30] predicted solar radiations at 12 different climatic location spread over India using ANN trained with Levenberg-Marquard (LM) algorithm. Sivamadhavi and Selvaraj [31] developed multilayer (3 and 4 layer) feed forward neural network (MLFFNN) to forecast mean monthly and daily global radiation in Tamilnadu, India. Notton *et al.* [32] developed ANN model to predict solar global radiation on inclined surfaces at the Mediterranean sites of Ajaccio in France. The trained ANN model was optimised and five year solar data were used for testing the model. Eissa *et al.* [33] presented a statistical model for predicting horizontal solar irradiance employing six thermal channels of the instrument, solar zenith angle, solar time, day number and eccentricity correction. However, it was suggested that adding more stations data during training phase could improve accuracy of the model.

ANN techniques have been used by many researchers as an alternative method for estimating solar radiation at different regions where IMD data is not available [34]. A comparative study pointed out that ANN based technique has an advantage over empirical regression based technique while predicting solar radiation at any location provided availability of sunshine hours data [35].

Few applications of genetic programming can be found in recent literatures mainly focused on predicting solar radiation. Landeras *et al.* [36] applied gene expression programming (GEP) for estimating daily incoming solar radiation and a comparative study has been done with ANN and ANFIS. Shavandi and Ramyani [37] employed the linear genetic programming (LGP) for estimating global solar radiation in Iran and it was claimed that the proposed model had a comparatively improved outcome with respect to the traditional Angstrom model.

## 3. Genetic programming for symbolic regression

Genetic Programming (GP) is an evolutionary algorithm used to automatically evolve computer programs [38] to perform a specific task. In essence it is an optimisation process which tries to find the optimal solution $s^*$ such that

$$s^* = \arg\min_{s \in S} f(s) \qquad (1)$$

where $S$ is the search space of the probable solutions and $f$ is a suitably defined fitness function. The optimal solution $s^* \in S$ minimizes $f$. In GP the candidate solutions $s \in S$ are functions of the form $s : \Gamma \to \Psi$ where $\Gamma$ and $\Psi$ are the spaces of the input and the output data of the programs from $S$. The strength of these evolutionary algorithms is that they do not try a brute force method for all the solutions in the input space $\Gamma$. Rather they mimic the process of evolution through a combination of operators simulating reproduction, crossover and mutation, giving rise to improved solutions over the iterations. In this paper the GP has been adapted for the purpose of symbolic regression to evolve a nonlinear function which fits a finite number of data points in the mean square sense [39]. Mathematically expressed, if there is an unknown function $h(x)$ then it is required to find another function $g(x)$ such that

$$h(x_i) = g(x_i) \;\forall x_i \in \Theta \qquad (2)$$

where, $\Theta$ is a set of samples taken from the interval of interest. In general $h(x)$ is not known precisely and only a set of sample values $\{(x, h(x)) \,|\, x \in \Theta\}$ is known. The advantage of using GP based symbolic regression over traditional linear and nonlinear regression models is that the structure of the model need not be specified *a priori* and the GP evolves both the structure and the parameters of the mathematical model. This has a huge advantage for the designer as it eliminates the need for speculative intuition on the part of the designer in choosing the appropriate regression model based on the nature of the data. In machine learning terminology, GP actually performs a kind of supervised learning. It is given a set of fitness cases or data points of the form $(x_i, y_i) \in \Gamma \times \Psi$, where $x_i$ represents one or more independent variables and $y_i$ is the output variable. The fitness function can then be expressed as

$$f(s) = \sum_i \| y_i - s(x_i) \| \tag{3}$$

where, $s(x_i)$ is the output of the GP evolved program $s$ for the input set $x_i$. $\| \bullet \|$ represents a metric like the Euclidean norm or 2-norm or the root mean square error on the output space $\Psi$ and $i$ is an iterator for all the given fitness cases.

Like other evolutionary algorithms (EAs), GP is based on the Darwinian principle of evolution and survival of the fittest. The solution variables are encoded in what is known as the genes or trees. At the start of the algorithm, the genes or expression trees are randomly initialized within the feasible space. Then they undergo reproduction, crossover and mutation to evolve fitter individuals in the succeeding generations. Crossover refers to the interchange of genetic material among the solutions. Mutation on the other hand refers to a random change within a gene itself. The crossover and mutation operations are stochastic ones and their probability of occurrence is pre-specified by the user. Unlike other EAs, in genetic programming based symbolic regression, the solutions or genes are encoded in the form of a tree as shown in Figure 1. The multi-gene GP or MGGP is used in this case for performing the symbolic regression. A multi-gene [3] is composed of one or more genes and each of these genes have a tree representation [3]. This helps in evolving simpler and more accurate functions [6]. Generally the maximum number of levels in a tree is confined to a specific small number to decrease the bloat in the solutions and also obtain fairly accurate results in a relatively short run-time [40]. The predicted output ($\hat{y}$) of each multi-gene regression model can be represented as:

$$\hat{y} = w_0 + \sum_{i=1}^{T} w_i \Phi_i \tag{4}$$

where, $w_0$ is the bias term, $T$ represent the number of trees, $w_i \; \forall i \in [1, T]$ are the gene or tree weights, $\Phi_i \; \forall i \in [1, T]$ represent the individual trees. Each tree is composed of a function of zero or more of the $N$ input variables $x_i \; \forall i \in [1, N]$.

Figure 1 is a tree representation of a multi-gene model with inputs $x_1, x_2$ and outputs $y$. The weights $w_0, w_1, w_2$ are obtained automatically by the MGGP algorithm by minimizing the mean squared error. The fittest genes are carried on to the next generation by directly copying them without any modifications. The other genes undergo crossover and mutation to form new individuals. Both these operations occur with the two-point high-level crossover operator. For the crossover operator the parent genes are selected from the population and two crossover points are randomly identified in the genes. The sub-trees formed at each gene of the crossover point are interchanged by swapping them between the two parents. This gives rise to two new child genes for the next generation. Since the depth of each tree has a specified maximum limit, the off-spring with genes greater than the specific depth have some of them randomly deleted to conform with the depth specifications.

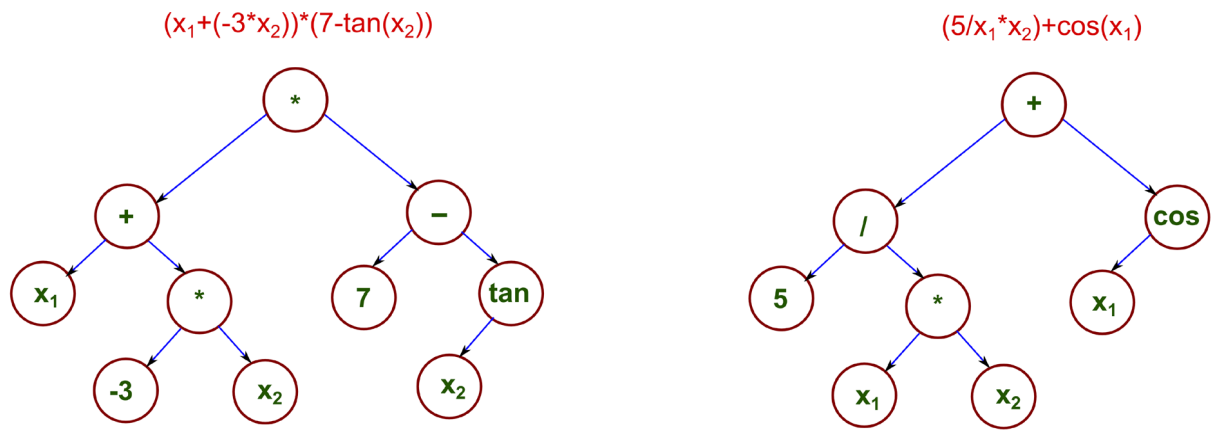

$$y = w_0 + w_1\{(x_1+(-3*x_2))*(7-\tan(x_2))\} + w_2\{(5/x_1*x_2)+\cos(x_1)\}$$

Figure 1: Tree representation of a MGGP model.

In Figure 2 the mutation operation is shown. A random node in the parent tree is selected and a randomly generated sub-tree replaces the original node and its sub-elements. Thus a new child expression is generated from the original symbolic expression. Figure 3 illustrates the crossover operation in GP based symbolic regression. Since the mathematical operators are encoded as trees to represent expressions, an interchange of some randomly selected sub-trees of the parent genes give rise to new child genes representing different expressions.

The pseudo-code of the GP algorithm is outlined next [38], [41].

```
Genetic Programming Algorithm
```
**Input :** A problem $\Lambda:[f,\Omega]$        //$f$-fitnessfunction,$\Omega$-instruction set
**Output :** A program to resolve $\Lambda$
**begin**
  $S \leftarrow \{s \leftarrow \texttt{InitializeRandomPopulation}(\Omega)\}$
  **repeat**
    **for** $s \in S$ **do**
      $store\_fitness \leftarrow f(s)$
    **end for**
    $S' \leftarrow \phi$
    **repeat**
      $(p_1, p_2) \leftarrow \texttt{TournamentSelection}(S)$
      $S' \leftarrow \texttt{Reproduction}(S)$
      $(c_1, c_2) \leftarrow \texttt{Crossover}(p_1, p_2)$
      $(c_1) \leftarrow \texttt{Mutation}(c_1, \Omega)$
      $(c_2) \leftarrow \texttt{Mutation}(c_2, \Omega)$
      $S' \leftarrow S' \cup \{c_1, c_2\}$
    **until** $|S'| = |S|$
    $S \leftarrow S'$
  **until** $\texttt{StoppingCondition}(S)$
  **return** $\arg\max_{s \in S} store\_fitness$
**end**

The plain lexicographic tournament selection as proposed in Luke and Panait [42] is used in the present GP algorithm. The lexicographic parsimony pressure is a simple mechanism for optimizing both fitness and tree size by treating them both as the primary and secondary objectives respectively in lexicographic ordering. The technique does not assign a new fitness value, but uses a modified tournament selection operator to take the size into account.

### 4. Estimation problem of global solar energy

The GP based algorithm utilizes latitude, altitude, longitude, months of the year, temperature ratio ($T/T_0$) and mean duration of sunshine per hour ($S/S_0$) for predicting clearness index ($H/H_0$). These input data are selected as a result of their relationships with global solar radiation [43]. The clearness index essentially reflects the availability of solar irradiance on the flat surface and changes with atmospheric conditions. Relative daily sunshine hours ($S/S_0$) define climatic condition of that location. The other variables like latitude, altitude and longitude indicate the geographical location and hence influence the solar radiation at that place. Also the time of the year reflects the seasonal alterations which have a significant effect on the solar irradiation. Hence these variables are chosen as inputs to the GP algorithm which generates a nonlinear combination of these variables to arrive at an analytical expression denoting the prediction formula. The mathematical relationship between these variables are introduced briefly next.

Daily maximum possible extraterrestrial sunshine hours and maximum extraterrestrial solar irradiance are calculated from the following equations [44]

$$S_0 = \left(\frac{2}{15}\right)\omega_s \tag{5}$$

$$H_0 = \frac{24 \times 3.6 \times 10^3 \times I_{sc}}{\pi}\left(1 + 0.033\cos\frac{360n}{365}\right) \times \left(\cos\phi\cos\delta\sin\omega_s + \frac{\pi\omega_s}{180}\sin\phi\sin\delta\right) \tag{6}$$

where

$$\delta = 23.45\sin\left(360\frac{284+n}{365}\right) \tag{7}$$

$$\omega_s = \cos^{-1}\left(-\tan\phi\tan\delta\right) \tag{8}$$

where, n is day of the month, $I_{sc}$ = 1367 Wm$^{-2}$ is the solar constant, $\phi$ is the latitude of location, $\delta$ is the declination angle and $\omega_s$ is the sunset hour angle.

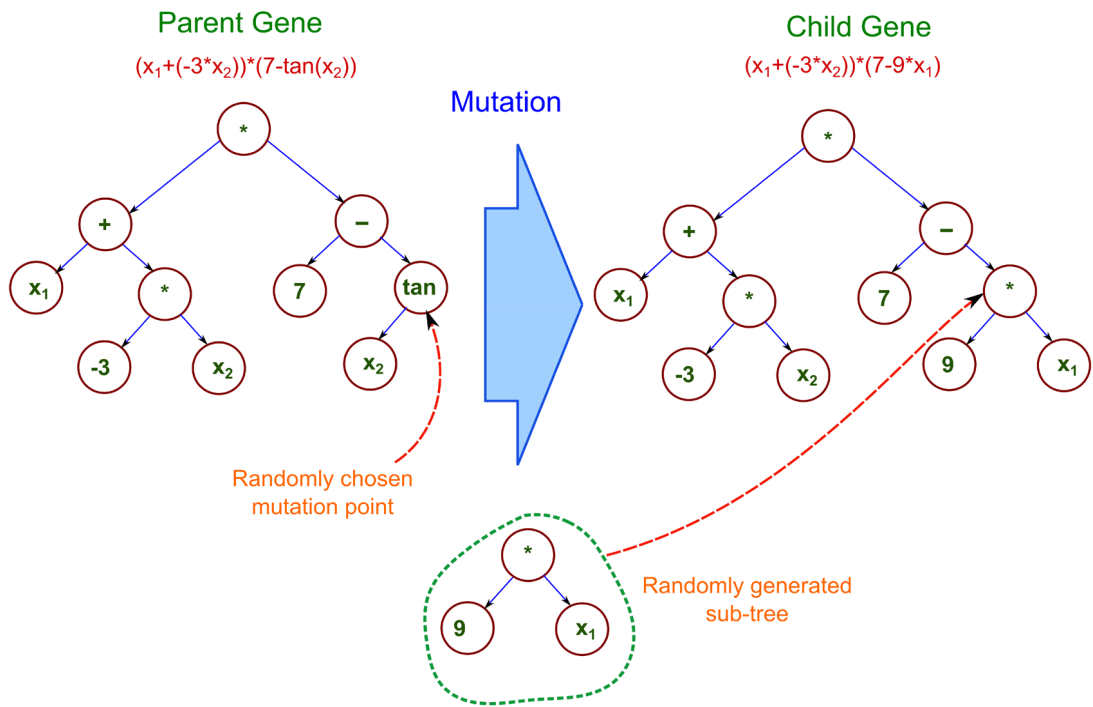

Figure 2: Mutation operation in Genetic Programming.

Angstrom [45] developed the most simple and extensively used model for predicting solar radiation which was later modified by Prescott [46]

$$\frac{H}{H_0} = a + b\left(\frac{S}{S_0}\right) \tag{9}$$

where, a and b are empirical constant connected to the locations; *H* represents daily global solar irradiation; $H_0$ represents daily global solar irradiation under clear sky conditions; *S* is the monthly mean of daily sunshine hours and $S_0$ is the maximum daily sunshine duration or day length. However this has poor prediction accuracy and hence models with more accurate prediction capabilities are required.

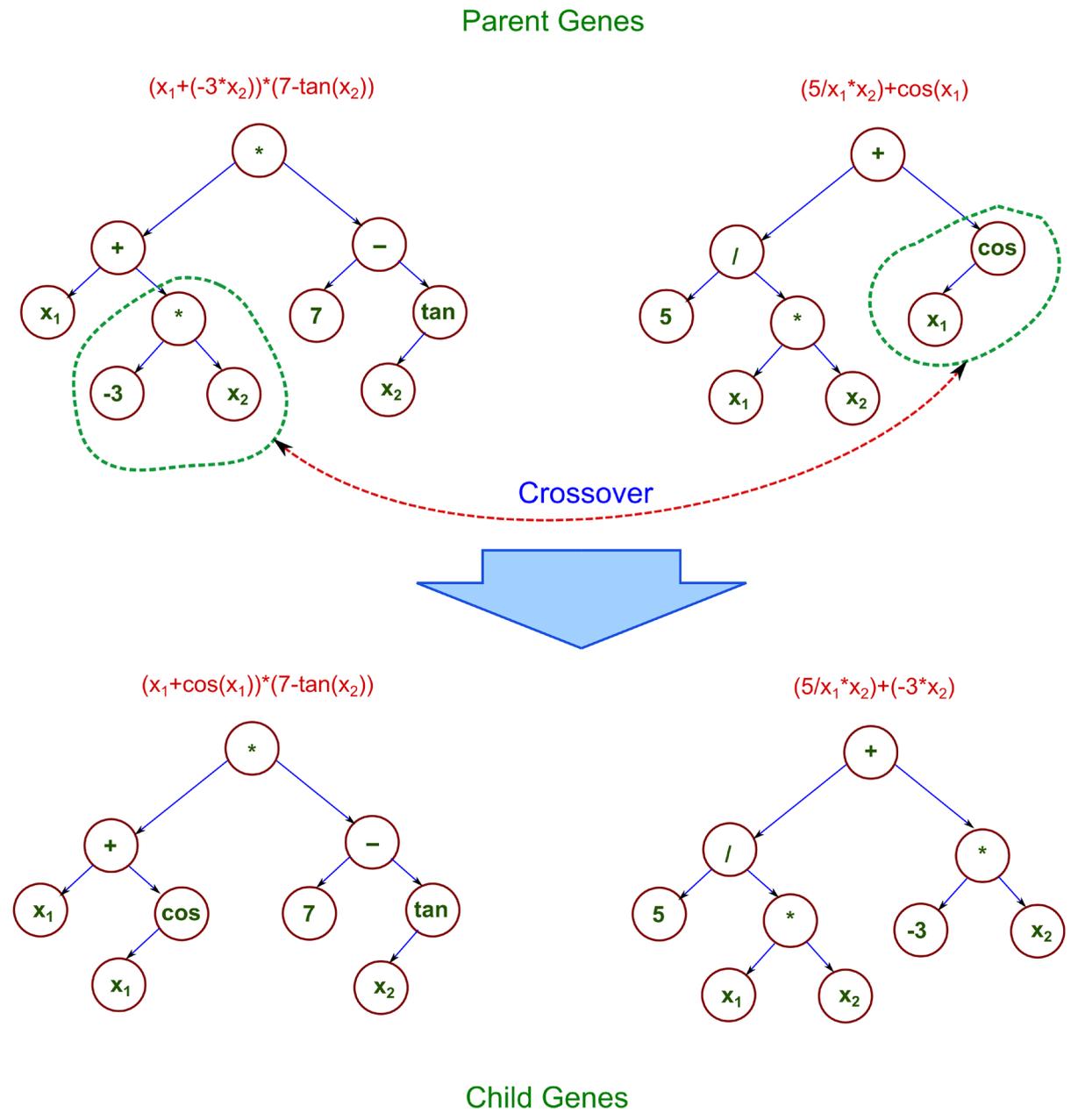

Figure 3: Crossover operation in Genetic Programming.

The input data encompasses distinct climatic regions and geographical stations in India and each station's data are attributed in Table 1. This study illustrates a common tactic on global solar irradiance estimation at any site in India using MGGP based symbolic regression approach. In principle, geographical and sunshine duration data is readily available at most of the measurement sites and vastly used for defining climatic condition of the site. Many researchers have used geographical coordinates such as latitude, longitude and altitude, meteorological data such as sunshine duration and mean temperature and the corresponding month as inputs for estimating solar potential [47][23][48]. However, effect of other atmospheric factors viz. aerosols, dust, moisture, wind velocity is negligible [49]. An attempt has been made to verify the effect of dust (air pollution) on the performance of PV solar panel. The study revealed that air pollution considerable deteriorate the energy yield up to 6.5% resulting an annual income loss of 40€/kW$_p$ i.e. 1% of turnkey specific price of domestic PV-generators [50].

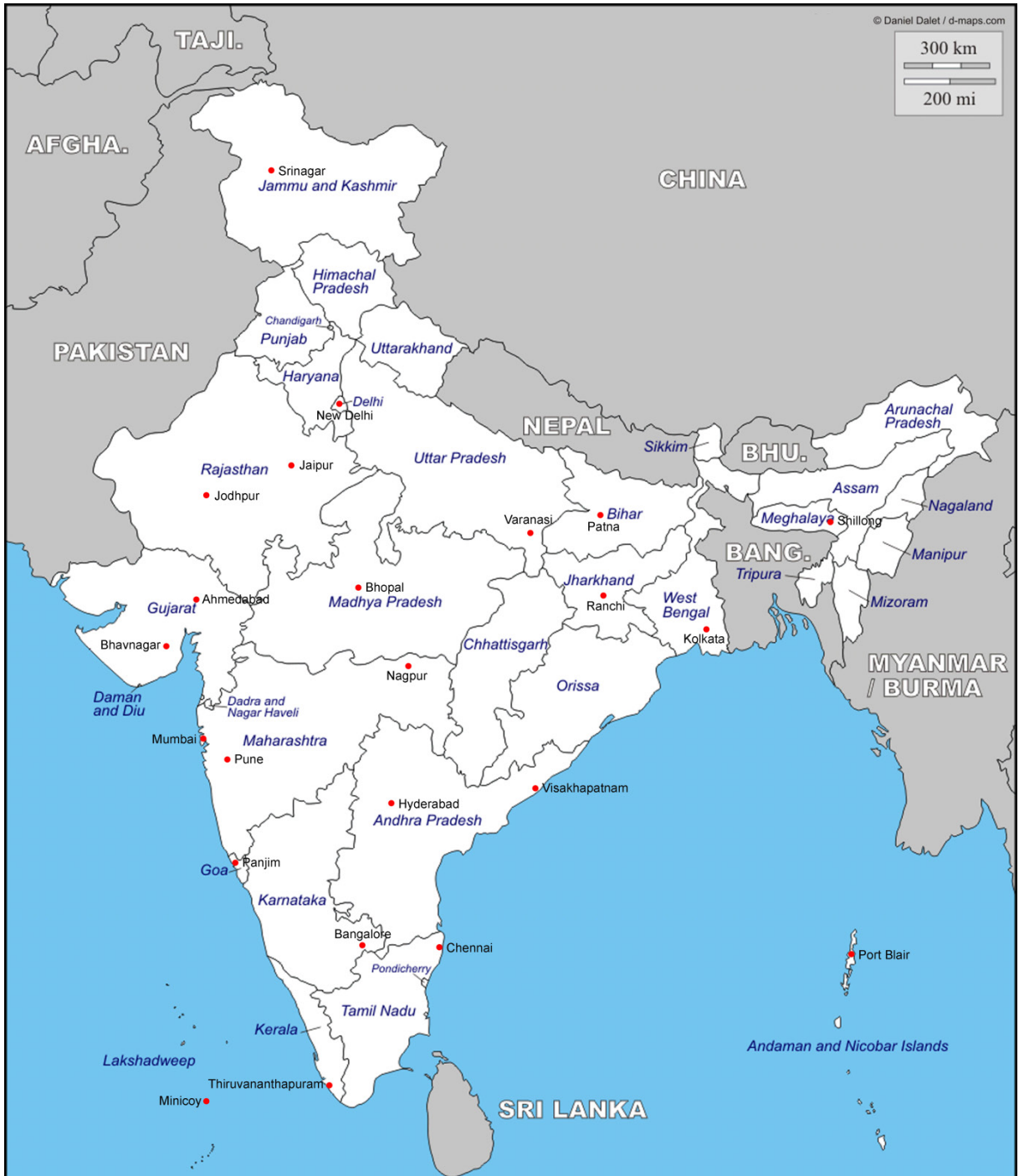

Figure 4: Location of different metrological stations considered in the paper [51] (Reproduced with permission from www.d-maps.com).

Table 1: Geographical locations of the considered stations

| Station | Latitude (°N) | Longitude (°E) | Height above the mean Sea Level (m) | Climate Zone |
|---|---|---|---|---|
| New Delhi | 28.58 | 77.2 | 216 | Composite |
| Nagpur | 21.1 | 79.05 | 31 | Composite |
| Ahmedabad | 23.07 | 72.63 | 55 | Hot & Dry |
| Jodhpur | 26.3 | 73.02 | 224 | Hot & Dry |
| Kolkata | 22.65 | 88.45 | 6 | Warm & Humid |
| Vishakhapatnam | 17.72 | 83.23 | 3 | Warm & Humid |
| Shillong | 25.57 | 91.88 | 1600 | Cold & Cloudy |
| Srinagar | 34.08 | 74.83 | 1586 | Cold & Cloudy |
| Jaipur | 26.92 | 75.98 | 431 | Hot & Dry |
| Varanasi | 25.33 | 83 | 80.71 | Composite |
| Patna | 25.6 | 85.12 | 53 | Composite |
| Bhopal | 23.26 | 77.4 | 427 | Composite |
| Ranchi | 23.35 | 85.33 | 629 | Composite |
| Bhavnagar | 21.77 | 72.15 | 24 | Hot & Dry |
| Mumbai | 18.96 | 72.82 | 11 | Warm & Humid |
| Pune | 18.54 | 73.86 | 560 | Hot and Dry |
| Hyderabad | 17.36 | 78.46 | 542 | Composite |
| Goa | 15.49 | 73.82 | 7 | Warm & Humid |
| Chennai | 13 | 80.18 | 16 | Warm & Humid |
| Bangalore | 12.96 | 77.58 | 921 | Moderate |
| Port Blair | 11.62 | 92.72 | 79 | Warm & Humid |
| Minicoy | 8.28 | 73.03 | 2 | Warm & Humid |
| Thiruvananthapuram | 8.5 | 76.9 | 10 | Warm & Humid |

Figure 4 shows the location of the different metrological stations on the Indian map. As can be seen, the locations include a lot of diversity and hence the model is expected to capture the distinct climatic features of each region and yet generalize well for the whole country. The Indian Metrological Data in [51] provides mean monthly global solar radiant exposure data for these radiation stations in India from a period of 1986-2000. The next section describes the results obtained from the MGGP simulation runs.

5. **Numerical simulation results with MGGP algorithm and discussions**

There are six input variables as latitude ($\Phi$), longitude ($\lambda$), altitude ($\alpha$), month of the year ($\eta$), $S/S_0$ and $T/T_0$ and are represented by the input vector $\hat{x} = [\Phi \ \lambda \ \alpha \ \eta \ S/S_0 \ T/T_0]$. The output variable ($y$) is $H/H_0$. The number of data-points in the database is 192. Different statistical measures for the whole data-set including input and output variables are shown in Table 2.

Table 2: Statistical measures of input and output variables of the whole data-set

| Statistical measure | Input variables | | | | | | Output variable |
|---|---|---|---|---|---|---|---|
| | latitude ($\varphi$) | longitude ($\lambda$) | altitude ($\alpha$) | month ($\eta$) | sunshine per hour ($S/S_0$) | temperature ratio ($T/T_0$) | clearness index ($H/H_0$) |
| Minimum | 8.280 | 72.630 | 2.000 | 1.00 | 0.460 | 0.026 | 0.256 |
| Maximum | 34.080 | 92.720 | 1586.000 | 12.00 | 1.002 | 0.940 | 0.698 |
| Mean | 19.924 | 77.826 | 317.875 | 6.500 | 0.769 | 0.714 | 0.556 |
| Standard deviation | 6.764 | 5.676 | 421.587 | 3.461 | 0.112 | 0.161 | 0.093 |

The solutions are obtained after scaling the original test data to zero mean and unit variance to overcome the bias of the different magnitudes of the variables on the final prediction. The mean and variance is calculated for the test data which is sampled randomly from the original dataset and consists of 70% of the values of the original dataset. The remaining 30% data is used for testing purposes. The training and testing datasets are not selected contiguously. The testing data is used for hold-out cross-validation, i.e. the model that is evolved by the GP technique based on the training data set is applied to the testing data set to check the accuracy of prediction on this untrained sample. The mean of the six input variables $\hat{x}_i \, \forall i \in \{1, 2, \cdots, 6\}$ and the output variable $y$ is denoted by the vectors $\mu_x$ and $\mu_y$ respectively and is given in Equations (10) and (11).

$$\mu_x = [20.1766 \quad 77.4217 \quad 347.1111 \quad 6.5259 \quad 0.7693 \quad 0.6982] \quad (10)$$

$$\mu_y = [0.5530] \quad (11)$$

The corresponding standard deviations of the variables are given by $\sigma_x$ and $\sigma_y$ in Equations (12) and (13).

$$\sigma_x = [6.9482 \quad 5.1183 \quad 445.5095 \quad 3.4959 \quad 0.1116 \quad 0.1707] \quad (12)$$

$$\sigma_y = [0.0933] \quad (13)$$

The scaling of the input variables ($\hat{x}_i$) is done to obtain the intermediate input variables ($x_i$) by using Equation (14)

$$x_i = \left(\hat{x}_i - \mu_{x_i}\right)/\sigma_{x_i} \quad \forall i \in \{1, 2, \cdots, 6\} \tag{14}$$

GP uses these intermediate inputs ($x_i$) to produce the intermediate output ($y$). To obtain the actual output variable ($\hat{y}$) from the intermediate output variable ($y$), Equation (15) is used.

$$y = \sigma_y \hat{y} + \mu_y \tag{15}$$

In the present simulation study, the MGGP algorithm is run with a population size of 100 and the maximum number of generations is set as 500. A tournament selection strategy is adopted for selecting the parent genes from the pool of available solutions. The tournament size is set to 3. The maximum depth of each tree in the multi-gene representation is set to 5. This allows some control over the complexity of the evolved expressions. The instruction set or the functions that are used for symbolic regression are $\left\{+, -, \times, \div, \sin, \cos, \sqrt{|\bullet|}, (\bullet)^2, \exp, \log|\bullet|\right\}$. The crossover probability is taken as 0.85, the mutation probability as 0.1 and the probability for direct reproduction is taken as 0.05. For simulation comparison we also ran the SGGP algorithm and the details of both GP variants are given in Table 3. For SGGP the number of trees T is set to one.

Figure 5 shows the final population of the GP run showing the trade-off between the accuracy of the fit and the complexity of the evolved models. The solutions points labelled in blue represent the set of dominated solutions while those in green represent the set of non-dominated solutions or the Pareto front. Solutions which are on the Pareto front are non-dominated in the sense that there are no solutions which have both a lower fitness and a lower model complexity simultaneously than these ones. In other words, if another solution has a lower fitness value then it must have a higher model complexity and vice-versa. From the solutions on the Pareto front, three solutions A, B and C as indicated by arrows are reported and their corresponding prediction accuracies and model complexities are compared. Figure 6 shows the convergence curve of the GP algorithm. The mean fitness of the overall population becomes almost constant after a particular number of generations. The fitness of the best individual changes marginally after 700 generations.

The MGGP convergence characteristics in Figure 6 also indicate that 1000 generations is sufficient for the convergence of the algorithm. It can be seen that the objective function does not change significantly near the end of the GP run. Also the associated curve of the mean fitness is plotted below it. It shows that the overall population loses diversity very quickly and running the GP algorithm for more number of generations is not going to yield a much better solution. However, it can also be seen that the algorithm should not be run for less than 100 generations for the present case, as the solution would not have converged sufficiently by then.

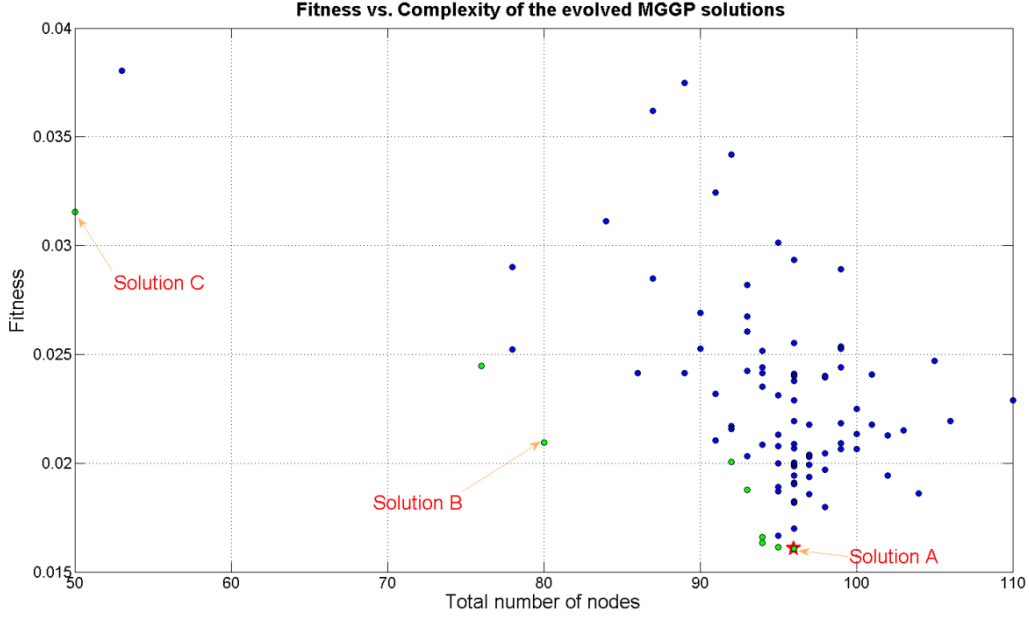

Figure 5: Fitness vs. complexity of the evolved MGGP solutions along with Pareto solutions.

Solution A in Figure 5 has the best fitness among all the solutions on the Pareto front, but has the highest model complexity (represented by the number of nodes in the GP tree). Solutions B and C have lower model complexities but have poor prediction performance. The evolved regression equation for Solution A (best fit) is given by

$$\frac{H}{H_0} = 0.4088\left(\frac{S}{S_0}\right) - 0.8869\left(\frac{T}{T_0}\right) + 0.4886\sin\left(\lambda - \frac{T}{T_0}\right) - 0.2464 e^{\sqrt{|\lambda\alpha|}} + 0.2872\sin(\phi\cos(\eta))$$

$$+ 0.2713\sin\left(\sqrt{\left|\sin\alpha + \cos\left(\frac{T}{T_0}\right) - e^{\eta}\right|}\right) - 0.2102\cos\left(e^{\log|3.599\eta|}\right) - 0.4267\left(e^{2\log(|\lambda\alpha|)}\right)$$

$$+ 0.4127\cos\left(\frac{S}{S_o}\right) + 1.791\cos\left(\frac{T}{T_0}\right) + 0.8566 e^{\left(\frac{T}{T_0}\right)} + 0.9245\sqrt{|\lambda|} + 0.2464\sin\left(\frac{S}{S_0}\right)$$

$$- 0.4886\sin\left(\frac{T}{T_0}\right) + \frac{0.009945\eta - 0.0358\frac{T}{T_0}}{\phi} + \frac{0.0008241\frac{T}{T_0}}{\log\left(\left|\frac{T}{T_0}\right/\phi\right|\right)} - 0.002966(\lambda)^2 \frac{S}{S_0} - 0.2441\frac{S}{S_0}\cos\left(\frac{T}{T_0}\right)$$

$$+ 0.1014\frac{T}{T_0}\cos\left(\frac{T}{T_0}\right) - \frac{0.1014\cos(\phi)(\phi + \lambda)}{\log\left(\left|\frac{\alpha}{\phi}\right|\right)} - 3.04$$

(16)

Table 3: Parameter settings for the two GP variants (MGGP and SGGP).

| GP algorithm parameters | Parameter settings |
|---|---|
| Population size | 100 |
| Number of generations | 1000 |
| Selection method | Plain Lexicographic tournament selection |
| Tournament size | 3 |
| Termination criteria | 1000 generations or fitness value less than 0.00001 whichever is earlier |
| Maximum depth of each tree | 5 |
| Maximum number of trees in an individual (for MGGP only) | 15 |
| Mathematical operators for symbolic regression | $\{+, -, \times, \div, \sin, \cos, \sqrt{|\cdot|}, (\cdot)^2, \exp, \log|\cdot|\}$ |

The regression equation for Solution B (relatively less complexity) as evolved by the GP algorithm is given by

$$\frac{H}{H_0} = 0.4105\left(\frac{S}{S_0}\right) + 0.4038\sin\left(\lambda - \frac{T}{T_0}\right) + 0.1659\sin(\phi\cos\eta) + 0.2612\sin\left(\begin{array}{c}\sqrt{|\sin\alpha|} \\ +\cos\left(\frac{T}{T_0}\right) - e^\eta\end{array}\right)$$

$$-0.2238\cos\left(e^{\log|(3.599\eta)|}\right) - 0.5035e^{2\log(|\lambda\alpha|)} + 0.2972\cos\left(\frac{S}{S_0}\right) + 0.7904\cos\left(\frac{T}{T_0}\right) + 0.7794\sqrt{|\lambda|}$$

$$-0.4038\sin\left(\frac{T}{T_0}\right) + \frac{0.0003371\left(\frac{T}{T_0}\right)}{\log(|\cos\eta|)} - 0.001213(\lambda)^2\left(\frac{S}{S_0}\right) - 0.02022\left(\frac{S}{S_0}\right)\cos\left(\frac{T}{T_0}\right)$$

$$+0.07645\left(\frac{T}{T_0}\right)\cos\left(\frac{T}{T_0}\right) - \frac{0.07645\cos(\phi)(\phi+\lambda)}{\log\left(\left|\frac{\alpha}{\phi}\right|\right)} - 1.05$$

(17)

The regression equation for Solution C (the simplest expression or lowest complexity) as evolved by the GP algorithm is given by

$$\frac{H}{H_0} = 0.6763\left(\frac{S}{S_0}\right) - 0.1884\left(\frac{T}{T_0}\right) + 0.1388\sin\left(\lambda - \frac{T}{T_0}\right) - 0.05235e^{\log|\lambda\alpha|}$$

$$-0.1668\cos\left(e^{\log|(3.599\eta)|}\right) + 0.07162\cos\left(\frac{S}{S_0}\right) + 0.9711\cos\left(\frac{T}{T_0}\right) + 0.05235\sin\left(\frac{S}{S_0}\right)$$

$$-0.1388\sin\left(\frac{T}{T_0}\right) + 0.05005\alpha\cos\left(\frac{T}{T_0}\right) - 0.1988\left(\frac{S}{S_0}\right)\cos\left(\frac{T}{T_0}\right) - \frac{0.05005\cos(\phi)(\phi+\lambda)}{\log\left|\frac{\alpha}{\phi}\right|} - 0.6498$$

(18)

Figure 7, Figure 9 and Figure 11 show the predicted versus the actual data for the training and the validation set for the three solutions A, B and C respectively. Figure 8, Figure 10 and Figure 12 show the root mean squared (RMS) error on the training and validation sets for the three solutions respectively. It can be seen that the expression given by solution A is able to explain more than 97% of the variability in the training data. In other words, the error in prediction for the training data set is less than 3%. The evolved model generalizes well to the validation data set as well. This set basically represents data that the algorithm has not been trained with. It can be found that for such unseen data, the analytical expression given by solution A in Equation (16) is able to predict the solar irradiation with over 90% accuracy.

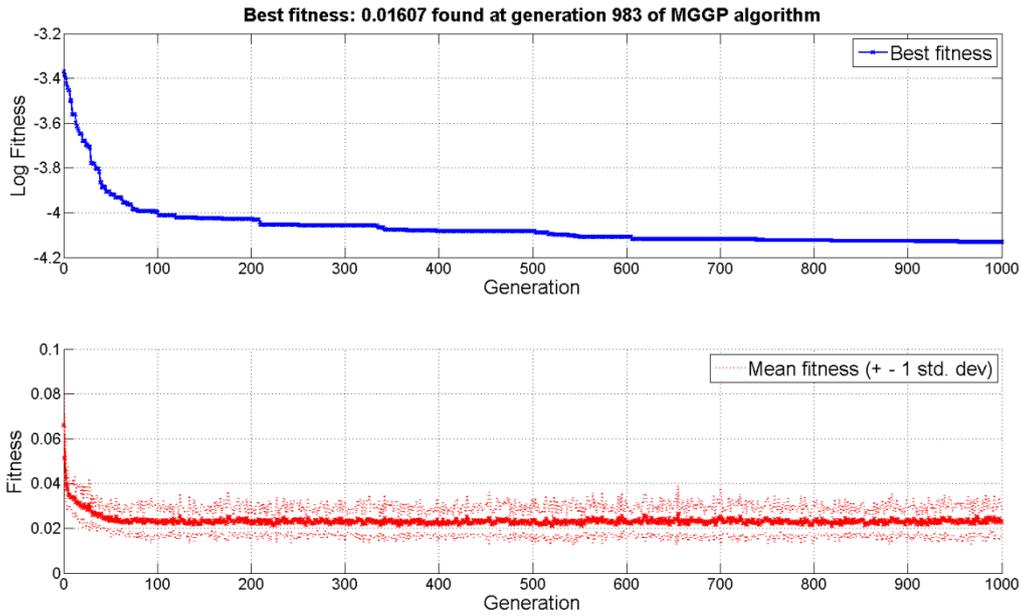

Figure 6: Convergence curves for the entire MGGP run with the best fitness and the mean fitness of the solutions.

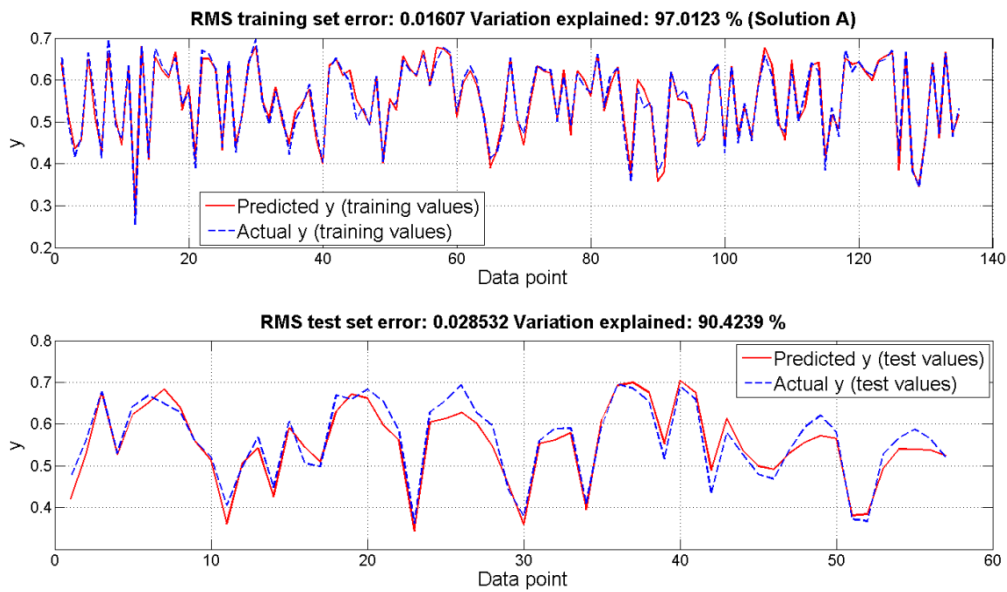

Figure 7: Prediction results of the best MGGP solution (Solution A) on the training and the testing dataset

Solution B and solution C on the other hand have less model complexity, i.e. they have fewer number of terms in the expression than Solution A, but their prediction capabilities are poorer as well. Model B still gives an impressive prediction accuracy of around 95% on the training data and more than 85% on the validation data set. Model C on the other hand gives a prediction accuracy of over 88% on the training set and over 79% on the validation set. This indicates that as the complexity of the prediction model decreases the percentage fit on the training data does not deteriorate drastically but the model does not generalize well to unseen data. In other words these models suffer from the problem of over-fitting the training data set. So essentially the choice of the model can be done based on the application at hand. If the model is used to predict only the data for the places that is already present in the training data set, then the lower complexity models like B or C may be used, since they give good prediction on the training data set. But if the model is to be used for predicting the values of solar irradiation at places that is not present in the present data set, then higher complexity models like A, must be used which generalize well to untrained data.

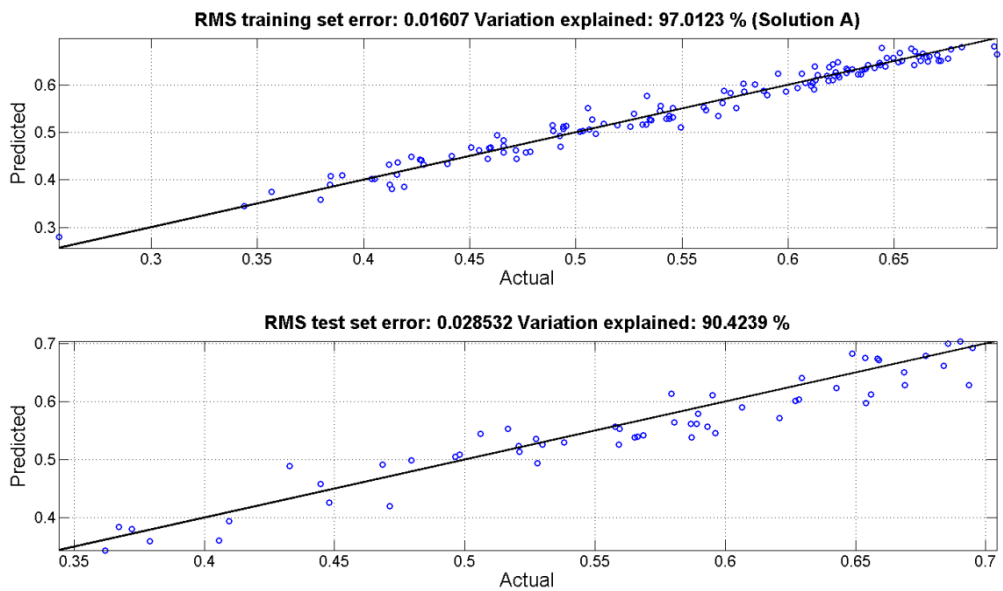

Figure 8: RMS error of the prediction with the best gene (Solution A) with MGGP on the training and test data

A comparison can be made with the models developed by other contemporary researchers to understand the effectiveness of this method. In [1], the same IMD data set has been used, as in the present paper to model the global solar irradiation using a generalized neural network (ANN) approach. An error of 4% is reported for the ANN case. A comparison is also made with the modelling with fuzzy logic approach and an error of 6% is reported. However this study has a few drawbacks. It essentially considers the whole set as the training data set and there is no validation data set. In other words the evolved models are not tested with unseen data. In our case with the GP based approach, less than 3% error is reported by the best solution A on the training data. Thus the MGGP method outperforms the other methods as reported by contemporary researchers on the training data set and also gives good prediction results for unseen data.

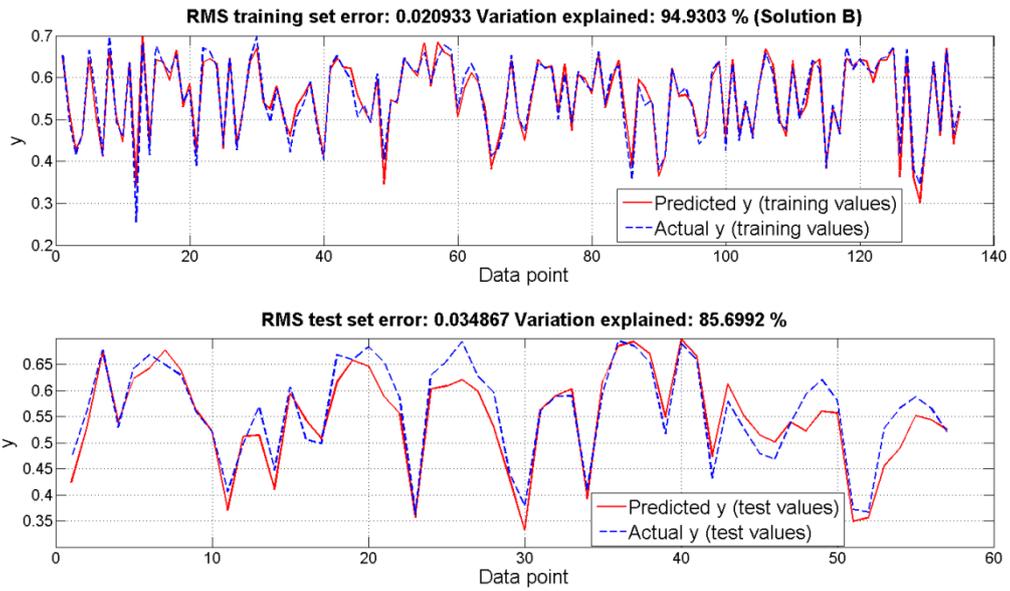

Figure 9: Prediction results of Solution B on the Pareto front of MGGP for the training and the testing dataset

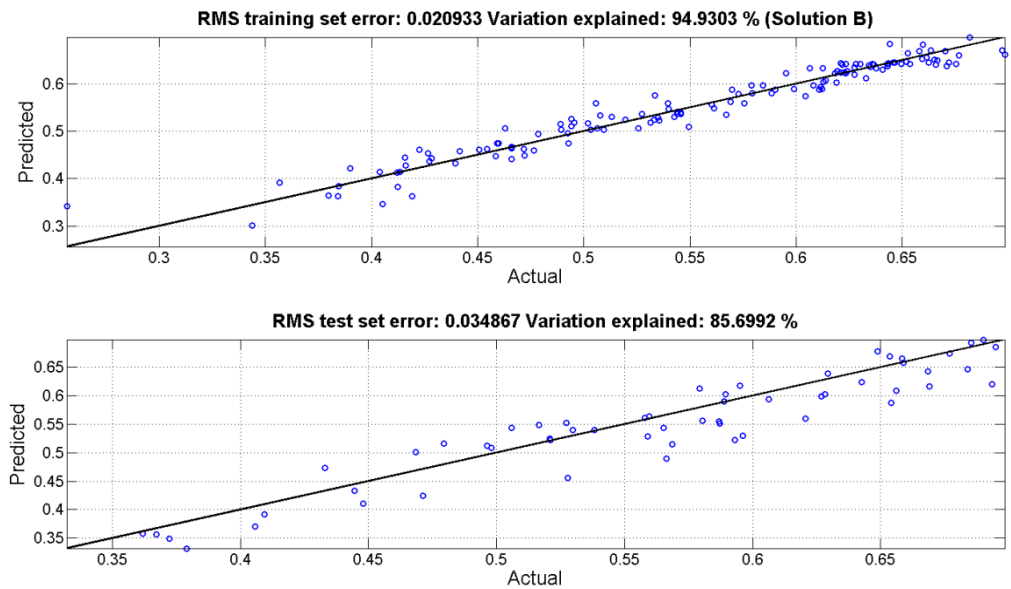

Figure 10: RMS error of the prediction with the solution B of MGGP on the training and test data

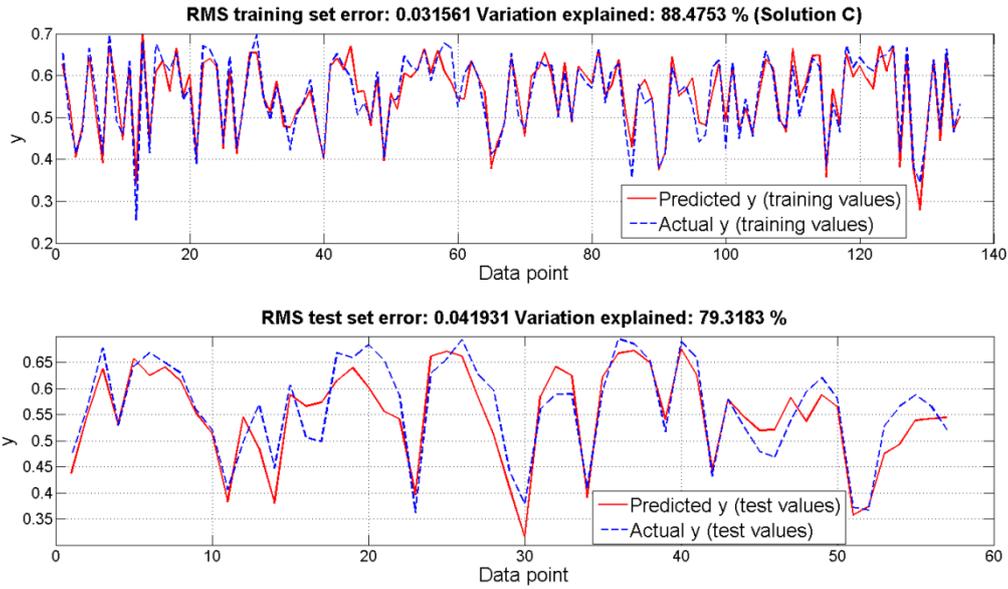

Figure 11: Prediction results of Solution C on the Pareto front of MGGP for the training and the testing dataset

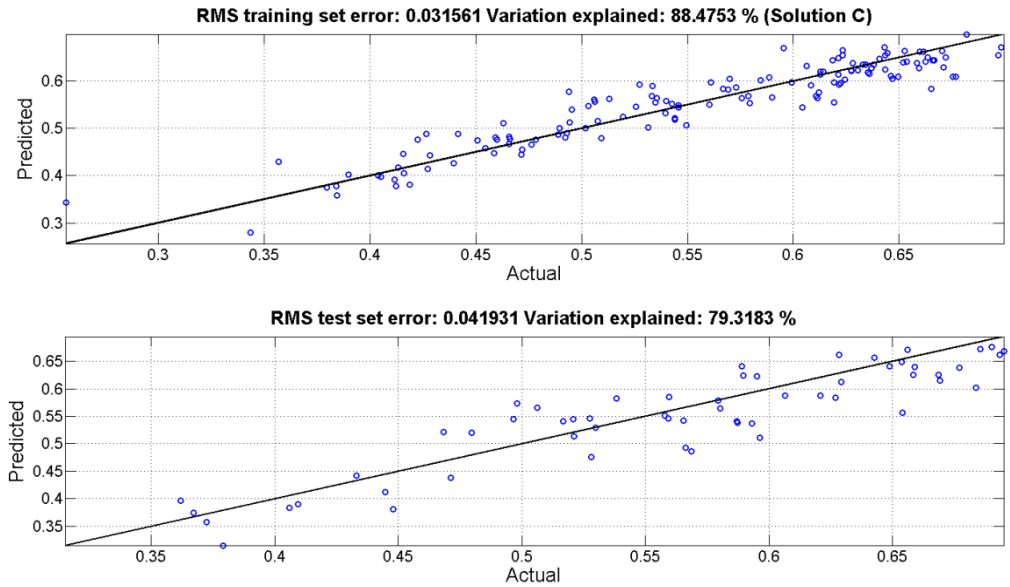

Figure 12: RMS error of the prediction with the solution C of MGGP on the training and test data

As a discussion it might be worth pointing out that the authors in [1] used a back propagation algorithm for training the neural networks. The back propagation algorithm is a gradient descent algorithm and hence it can get stuck in local minima. Hence global optimisation algorithms like swarm or evolutionary (S&E) optimisation can be used to train the ANNs to improve their prediction accuracies. It might be possible to obtain a higher accuracy with the ANN based methods than the GP ones, provided a proper structure, number of layers, activation functions etc. are chosen appropriately and the ANN is trained with global optimisation algorithms. However, even with a better fit, the ANN based models cannot give analytical expressions like the GP ones due to its inherent nature and design.

Since multiple models are evolved during a single GP run, hence the design engineer has an option to choose to make a trade-off from the various available models depending on their complexity and accuracy. Thus in cases where very high accuracy is not of prime concern, the design engineer can go for a low complexity model which is easier to calculate and can also shed insight on the most important variables affecting the prediction. The only disadvantage of the GP based symbolic regression is due to bloat where the complexity of the expression increases very fast with only a small increase in the fitness of the solutions. This results in redundant expressions which do not have significant effect on the final outcome but make the expression complicated and creates difficulties in discerning logical cause-effect relationships between the input data and the predicted outputs. There are various techniques for bloat control in GP and is still an active area of research [40].

The other advantage of GP over neural network or fuzzy models is that there is an evolutionary global optimisation algorithm inherent in the methodology itself. For fuzzy models, an additional optimisation algorithm is required to tune the fuzzy membership functions, rule bases etc. For neural networks, back-propagation or other gradient descent algorithms are generally required to adjust the weight of the individual nodes of the network to train it for a particular data set. Moreover, the choices of the number of hidden layers and number of nodes in each layer have to be chosen by trial and error. Thus fuzzy or neural network models need an additional optimisation mechanism for effective training of the model which is not required by GP as the evolutionary algorithm is embedded in it.

## 6. Statistical analyses of the MGGP algorithm and its comparison with SGGP and classical regression analysis

In this section, the result obtained using the MGGP algorithm is compared with a simpler technique SGGP and the classical regression techniques. In the previous section, it is already mentioned that the inherent intelligent optimization routine present in both version of the GP algorithm helps to avoid local minima like the gradient based back-propagation training of neural networks. It is known that the back propagation neural network training often gets trapped in local minima since it relies on the gradient of error [52]. On the other hand, GP is based on the evolution of a population which exchanges their information in an intelligent way using crossover and mutation. Like other evolutionary algorithms, these characteristics give GP the power to get out of local minima and converge towards the global one. The other option is to train the neural networks with evolutionary algorithms instead of back propagation [53], so that the effect of convergence to local minima is ameliorated to a certain extent. However in such a case, the ANN might give a better fit, but results in no analytical expression like GP.

Also, in order to show consistency of the GP based data-fitting approach, statistical analyses of the MGGP and SGGP algorithms for 30 independent runs are reported in Table 4.

It shows that the MGGP algorithm consistently converges to a better fit than that with SGGP which is evident from the mean and standard deviation of the best solutions in Table 4.

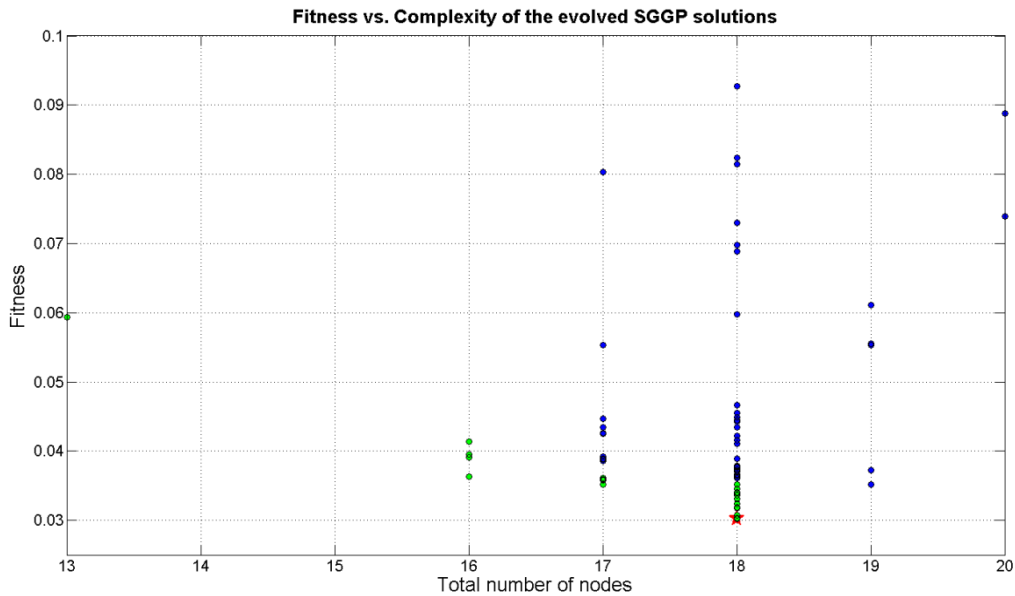

Figure 13: Fitness vs. complexity of the evolved SGGP solutions along with Pareto solutions.

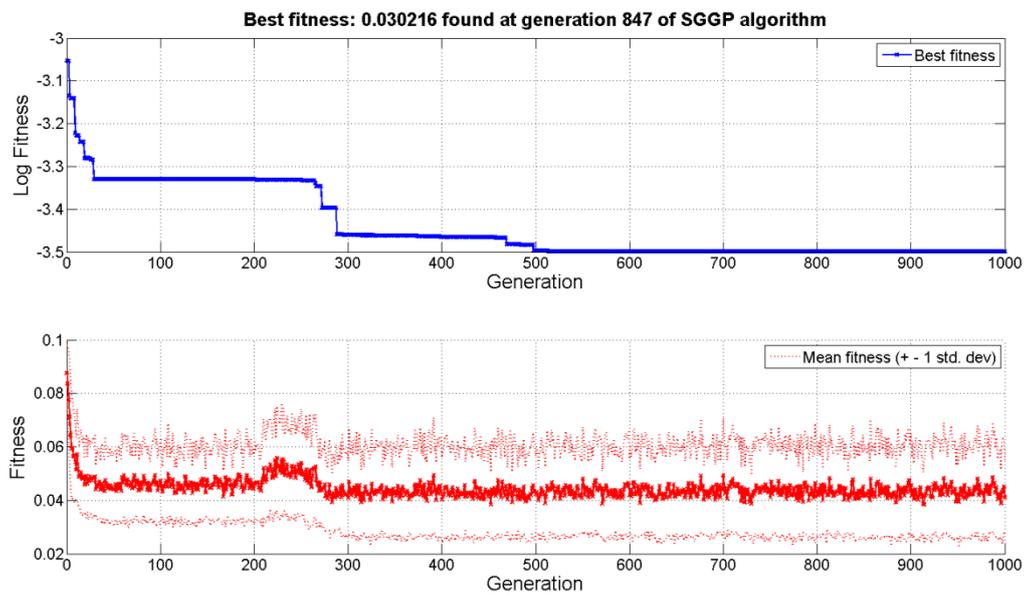

Figure 14: Convergence curves for the entire SGGP run with the best fitness and the mean fitness of the solutions.

Figure 13 shows the Pareto front of between the fitness vs. complexity for the SGGP algorithm. It is evident that for the best obtained solution in Figure 13 for the SGGP algorithm, the percentage fit and complexity both are significantly less than that using the MGGP algorithm in Figure 5. The convergence characteristics of the SGGP algorithm is also shown in Figure 14 which is also higher in this case compared to the MGGP one in Figure 6 in terms of best and mean fitness.

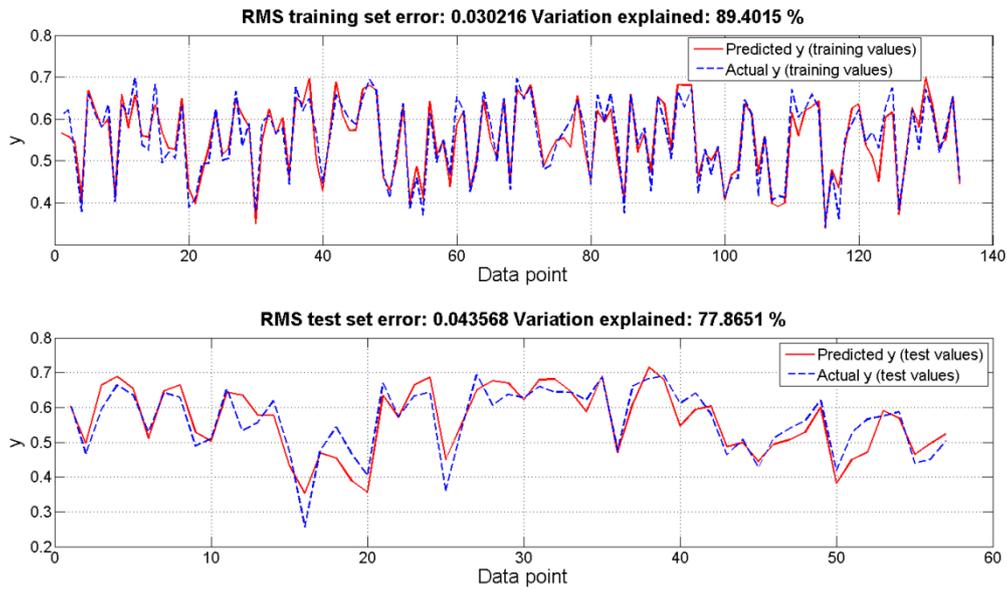

Figure 15: Prediction results of the best SGGP solution on the training and the testing dataset

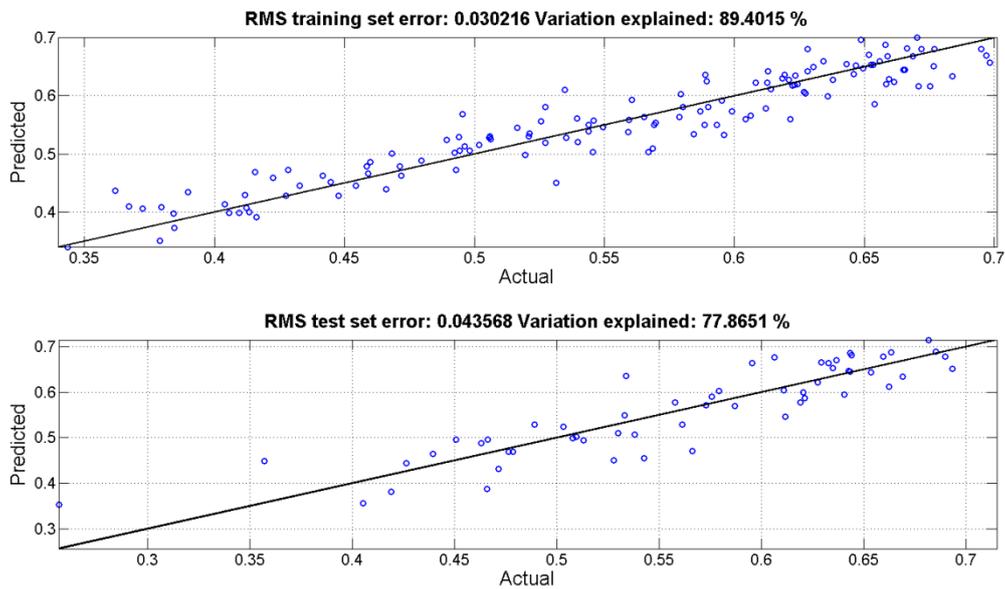

Figure 16: RMS error of the prediction with the best gene with SGGP on the training and test data.

The prediction results of the best SGGP based solution on the training and testing data-sets have been show in Figure 15 and Figure 16 which shows that the percentage fits are not so good than solution A and B of the MGGP case. Though the SGGP based result is not so good compared to the MGGP case, it is reported in equation (20) due to having less complexity.

$$\frac{H}{H_0} = 0.5631\left(\frac{S}{S_0} - \sqrt{|-\eta - 0.9726|} + \cos\left(\frac{T}{T_0}\right) - \sqrt{|\alpha|} - \sin\left(\frac{T}{T_0}\right)\right) + 0.7163 \quad (19)$$

Table 4: Statistics of the best solution for 30 independent runs of two GP variants

| Algorithm | Mean (μ) | standard deviation (σ) | max | min |
|---|---|---|---|---|
| MGGP | 0.017 | 0.001175022 | 0.0202 | 0.01607 |
| SGGP | 0.037 | 0.004155452 | 0.0493 | 0.030216 |

It has been already shown in the literature review in section 2 that neural network based predictions are good for complex data-sets. Here we apply multi-gene GP in order achieve the same goal, but which gives analytical expressions unlike ANN. It is well known that a classical regression approach can fit polynomial type models without significant nonlinearity like an ANN which would be an over-simplification towards the target and will yield significantly high RMSE. To demonstrate this point, here we have shown simulations with different regression models, which give poor results in comparison with the proposed MGGP approach. Therefore besides comparing the MGGP and SGGP fitting approach, we have also tried a much simpler approach of fitting the data with classical regression analysis with input vector $x = [\Phi \ \lambda \ \alpha \ \eta \ S/S_0 \ T/T_0]$ and output variable ($y$) as $H/H_0$. We have used four different cases like linear interactions, pure quadratic and quadratic terms in the regression model as tabulated in Table 5, along with mention about the fitting error (in terms of RMSE) and coefficient of determination ($R^2$), where $R^2$ is given by (21) and represents a good fit for its value closer to one i.e. very small sum of squares for the residuals.

$$R^2 = 1 - \frac{\sum_{i=1}^{n}(y_i - \hat{y}_i)^2}{\sum_{i=1}^{n}(y_i - \overline{y})^2}, \quad \overline{y} = \frac{1}{n}\sum_{i=1}^{n} y_i \quad (20)$$

Table 5: Regression models, their characteristics and corresponding RMSE and $R^2$ value

| Model | Characteristics of the model | RMSE | $R^2$ |
|---|---|---|---|
| Linear | intercept and linear terms | 0.0472 | 0.751 |
| Interactions | intercept, linear terms and all products of pairs of distinct predictors | 0.0326 | 0.891 |
| Pure quadratic | intercept, linear terms, and squared terms | 0.0362 | 0.858 |
| Quadratic | intercept, linear terms, interactions, and squared terms | 0.0307 | 0.907 |

The linear regression analysis have been run on the whole dataset by considering it as the training dataset and the RMSE is 0.0472 which is higher than both the testing and the training dataset results obtained in the GP runs. Amongst different combinations for the input variables the best solution by regression analysis was found using the quadratic model containing an intercept, linear terms, interactions, and squared terms with a RMSE of 0.0307 and $R^2$ of 0.907. On contrary, it has already been shown that the best MGGP has an RMSE of 0.01607 on the training dataset and 0.028532 on the testing dataset and also the SGGP has an RMSE of 0.030216 on the training data-set which are less than the conventional regression results. If the data was divided into training and testing, then the results for the regression would have been worse which has not been reported in the present context. The model structures for four different regression models with linear, interactions, pure quadratic and quadratic terms are shown in equation (22) in terms of the intercept ("1") and the input variables ($x_1 - x_6$) where each terms also contains a co-efficient that is estimated by the data-fitting process.

$$\begin{aligned}
y_{\text{linear}} &\sim 1 + x_1 + x_2 + x_3 + x_4 + x_5 + x_6 \\
y_{\text{interactions}} &\sim 1 + x_1 + x_2 + x_3 + x_4 + x_5 + x_6 + x_1 x_2 + x_1 x_3 + x_1 x_4 + x_1 x_5 + x_1 x_6 \\
&\quad + x_2 x_3 + x_2 x_4 + x_2 x_5 + x_2 x_6 + x_3 x_4 + x_3 x_5 + x_3 x_6 + x_4 x_5 + x_4 x_6 + x_5 x_6 \\
y_{\text{purequadratic}} &\sim 1 + x_1 + x_2 + x_3 + x_4 + x_5 + x_6 + x_1^2 + x_2^2 + x_3^2 + x_4^2 + x_5^2 + x_6^2 \\
y_{\text{quadratic}} &\sim 1 + x_1 + x_2 + x_3 + x_4 + x_5 + x_6 + x_1 x_2 + x_1 x_3 + x_1 x_4 + x_1 x_5 + x_1 x_6 \\
&\quad + x_2 x_3 + x_2 x_4 + x_2 x_5 + x_2 x_6 + x_3 x_4 + x_3 x_5 + x_3 x_6 + x_4 x_5 + x_4 x_6 + x_5 x_6 \\
&\quad + x_1^2 + x_2^2 + x_3^2 + x_4^2 + x_5^2 + x_6^2
\end{aligned} \quad (21)$$

Therefore as an effective model comparison, the RMSE for all the models i.e. SGGP, MGGP, linear regression have been reported in a Table 4 and Table 5.

## 7. Conclusion

In this paper, MGGP was employed to predict the total amount of measured solar irradiation from six different independent variables. The proposed methodology is backed by adequate numerical simulation and is proved to give better results than the previous approaches by other researchers using fuzzy logic and neural networks. Since it is a highly nonlinear and flexible regression method, the advantage of this method over the multiple regression method is that it is able to properly predict variables with high accuracy which are classified as outliers in the latter. Neural networks also have the same advantage, but GP scores over NN in that they offer analytical expressions and not black box models. Since analytical expressions are obtained by this methodology, it can help the analyst in improving the understanding of the inter-relationship between the various variables and the forecast. Since the set of instructions are transparent and easily interpretable by humans, it is possible to gain some novel understanding and insight into the underlying event that drives the process. Also for practising and field engineers this is advantageous, since the solar irradiation can be calculated using a simple hand calculator and no other sophisticated software is required. The only disadvantage of this method is due to bloat which results in an increase in the complexity of the evolved expression without significantly increasing the prediction accuracy. Analytical expressions obtained using MGGP approach was also compared with SGGP and four classical regression models which show that MGGP clearly outperforms the other two approaches. Future research may be directed towards extending the proposed solar irradiation prediction modelling approach by encompassing other several countries around the globe.